\documentclass{article} 

\usepackage{iclr2016_conference,times}

\usepackage{times}
\usepackage{relsize}	
\usepackage[T1]{fontenc} 
\usepackage[english]{babel}

\usepackage{epsfig}
\usepackage{graphicx}
\usepackage{wrapfig}
\usepackage[belowskip=0pt,aboveskip=0pt,font=small]{caption}
\usepackage[belowskip=0pt,aboveskip=0pt,font=small]{subcaption}
\usepackage{amsmath, amsthm, amssymb}
\usepackage{textcomp}
\usepackage{stmaryrd}
\usepackage{upgreek}
\usepackage{bm}
\usepackage{cases}
\usepackage{mathtools}
\usepackage{color}

\usepackage[pagebackref=true,breaklinks=true,letterpaper=true,colorlinks,bookmarks=false,citecolor=green]{hyperref}

\usepackage{algorithm}
\usepackage{algpseudocode}

\usepackage{multirow}
\usepackage{rotating}
\usepackage{booktabs}
\usepackage{tabularx}

\usepackage{enumitem}
\usepackage[olditem,oldenum]{paralist}

\usepackage{alltt}
\usepackage{listings}

\belowdisplayskip \abovedisplayskip

\newlength{\sectionReduceTop}
\newlength{\sectionReduceBot}
\newlength{\subsectionReduceTop}
\newlength{\subsectionReduceBot}
\newlength{\abstractReduceTop}
\newlength{\abstractReduceBot}
\newlength{\captionReduceTop}
\newlength{\captionReduceBot}
\newlength{\subsubsectionReduceTop}
\newlength{\subsubsectionReduceBot}

\newlength{\eqnReduceTop}
\newlength{\eqnReduceBot}

\newlength{\horSkip}
\newlength{\verSkip}

\newlength{\figureHeight}
\usepackage{mysymbols}

\usepackage{url}
\usepackage{xspace}
\usepackage{comment}
\usepackage{color}
\usepackage{afterpage}
\usepackage{pdfpages}
\usepackage{framed}
\usepackage{fancybox}
\usepackage[normalem]{ulem}

\title{Reducing Overfitting in Deep Networks by Decorrelating Representations}

\author{Michael Cogswell \\
Virginia Tech \\
Blacksburg, VA \\
\texttt{cogswell@vt.edu} 
\And
Faruk Ahmed \\
Universit\'e de Montr\'eal \\
Montr\'eal, Quebec, Canada \\
\texttt{faruk.ahmed@umontreal.ca} 
\And
Ross Girshick \\
Facebook AI Research (FAIR) \\
Seattle, WA \\
\texttt{rbg@fb.com} 
\And
Larry Zitnick \\
Microsoft Research \\
Seattle, WA \\
\texttt{larryz@microsoft.com} 
\And
Dhruv Batra \\
Virginia Tech \\
Blacksburg, VA \\
\texttt{dbatra@vt.edu}
}

\iclrfinalcopy 

\begin{document}
\maketitle
\begin{abstract}
One major challenge in training Deep Neural Networks is preventing overfitting.
Many techniques such as data augmentation and novel regularizers such as Dropout
have been proposed to prevent overfitting without requiring a massive amount
of training data. 
In this work, we propose a new regularizer called DeCov which leads to significantly
reduced overfitting (as indicated by the difference between train and val performance),
and better generalization. Our regularizer encourages diverse or non-redundant
representations in Deep Neural Networks by minimizing the cross-covariance of
hidden activations. This simple intuition has been explored in a number of past
works but surprisingly has never been applied as a regularizer in supervised
learning. Experiments across a range of datasets and
network architectures show that this loss always reduces overfitting
while almost always maintaining or increasing generalization performance
and often improving performance over Dropout.
\end{abstract}


\section{Introduction}
\label{sec:intro}

Deep Neural Networks (DNNs) have recently achieved remarkable success on
a wide range of tasks -- \eg, image classification on ImageNet~\citep{krizhevsky_nips12},
scene recognition on MIT Places~\citep{zhou_nips14}, image captioning with MS COCO~\citep{coco, show_and_tell, chen2015mind},
and visual question answering~\citep{VQA}.
One significant reason for improvement of these methods over their predecessors has to do with scale.
Faster computers coupled with optimization improvements such Batch Normalization, Adaptive SGD, and
ReLus let us quickly train wider and deep networks. Access to large annotated datasets and
regularizers such as Dropout has provided significant reduction in the amount of
overfitting in these large networks,
thus enabling the performance we see today.

In this paper, we focus on the problem of overfitting, which is observed when a high capacity model (such as a DNN)
performs very well on training data but poorly on held out data.
Even when trained on large annotated datasets (such as ImageNet~\citep{imagenet} or Places~\citep{zhou_nips14}, containing
millions of labelled images), deep networks are susceptible to overfitting. This problem
is further exacerbated when moving to new domains and tasks -- since DNNs tend not to
generalize with a few examples, each new task tends to require curating and annotating
a new large dataset. While there has been some success with transfer learning~\citep{girshick2014rcnn, donahue_icml14, yosinski2014transferable},
networks still overfit.

A promising alternative to creating even larger datasets is to apply different forms of regularization 
to the network while training to avoid overfitting. 
These methods include regularizing the norm of the weights~\citep{tikhonov_regular}, 
Lasso~\citep{tibshirani_lasso}, Dropout~\citep{dropout}, DropConnect~\citep{wan2013dropconnect},
Maxout~\citep{goodfellow2013maxout}, etc. 

One particular regularizer of interest to DNNs is Dropout~\citep{dropout}, 
which attempts to prevent co-adaptation of neuron activations. 
Co-adaptation occurs when two or more hidden units rely on one another to perform some function
which helps fit training data, thus becoming highly correlated.
Co-adaptation is reduced by Dropout using an approximate model averaging technique 
that sets a randomly selected set of activations to zero at training time.  
\cite{dropout} show that this has a regularizing effect, leading
to increased generalization and sparser, less correlated features. Notice that this
is without \emph{explicitly} encouraging decorrelation in hidden activations.

\begin{figure}[!h]
    \centering
    \vspace{-15pt}
	\begin{subfigure}[b]{0.49\textwidth}
        \vspace{10pt}
        \includegraphics[width=\columnwidth]{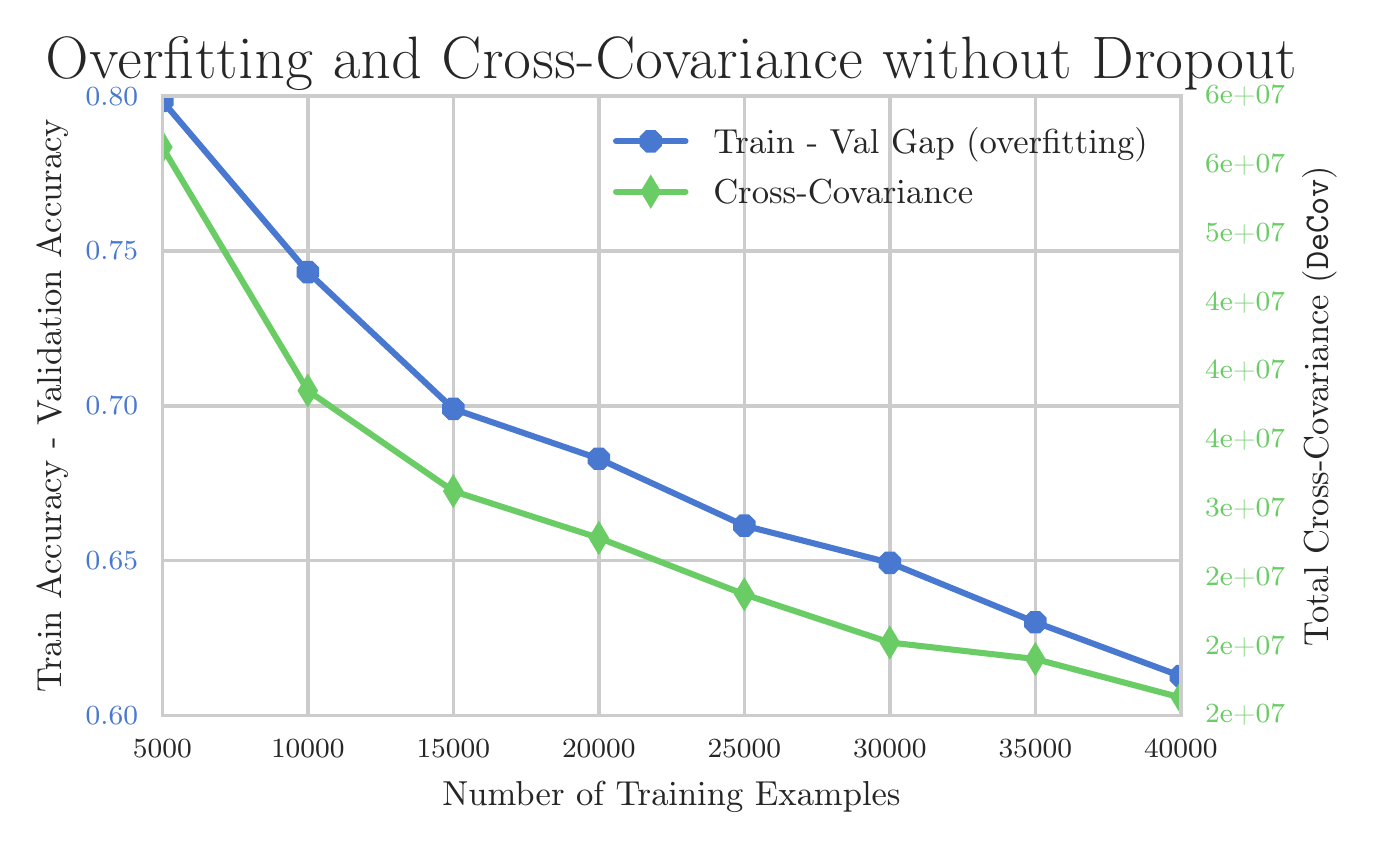}
        \caption{Without Dropout}
        \label{fig:teaser_nodrop}
    \end{subfigure}
	\begin{subfigure}[b]{0.49\textwidth}
        \vspace{10pt}
        \includegraphics[width=\columnwidth]{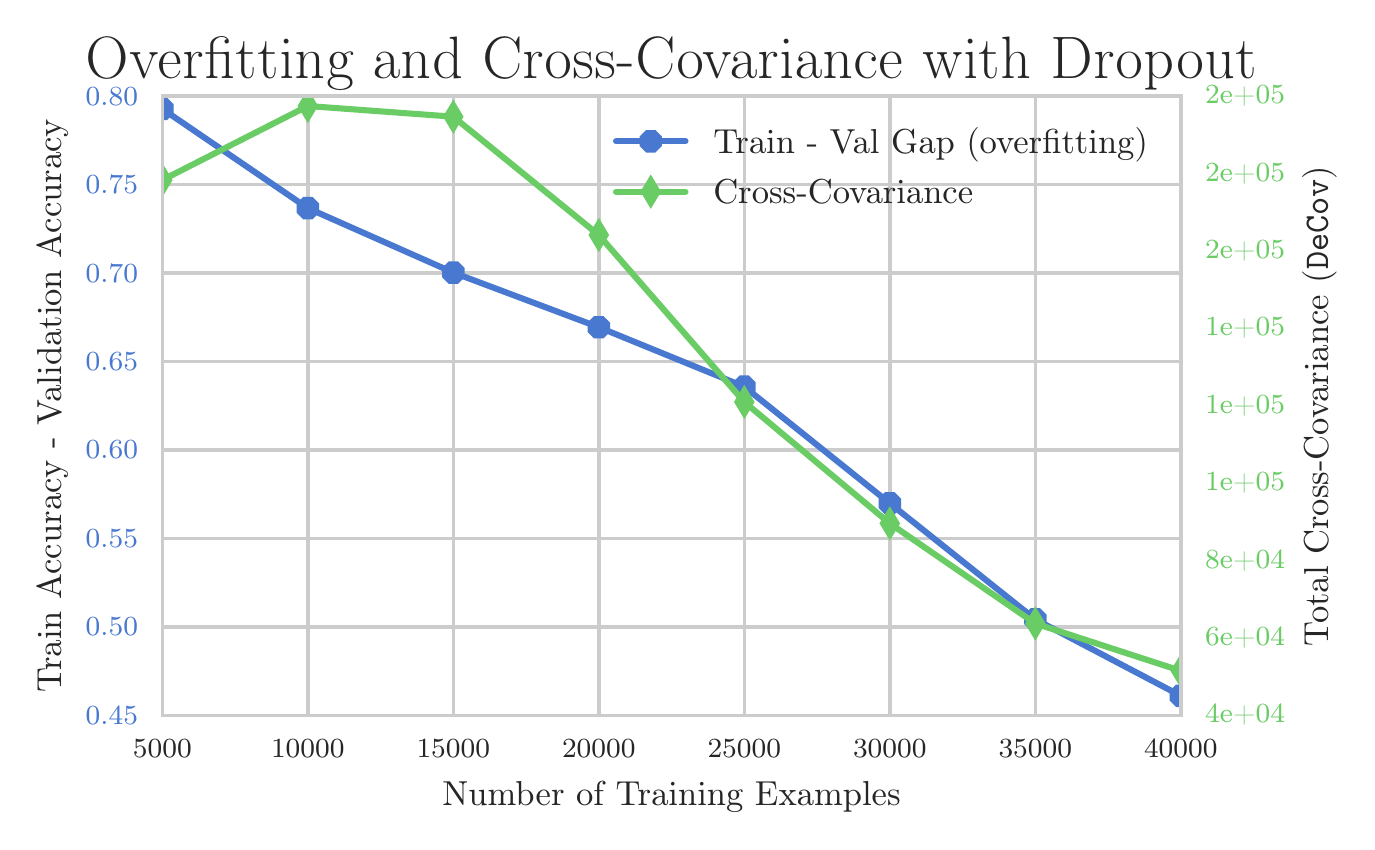}
        \caption{With Dropout}
        \label{fig:teaser_drop}
    \end{subfigure}
    \caption{Two principal ways to prevent overfitting in deep models
        are to train with more data (x axis) and to train with Dropout (right plot).
        As expected, both of these decrease validation error (left axis), but they
        also happen to decrease hidden activation cross-covariance (right axis).
        We investigate whether explicitly minimizing cross-covariance can lead to reduced overfitting. }
    \label{fig:teaser}
\end{figure}

To further investigate the relationship between hidden activation correlations and overfitting,
we show in \figref{fig:teaser} two quantities from a CNN
trained for image classification on CIFAR100~\citep{cifar10} -- (1) the
amount of overfitting in the model (as measured by the gap between train and val accuracy),
and (2) the amount of correlation in hidden activations (as measured by the Frobenius
norm of the sample cross-covariance matrix computed from vectors of hidden activations;
details in \secref{sec:approach}). Both these quantities of interest are reported as a function
of amount of training data (x-axis) and with/without Dropout (left/right subplot). As
expected, both increased training data and Dropout have a regularizing
effect and lead to reduced overfitting. 

The figure also shows an interesting novel trend -- as the amount of overfitting reduces, so
does the degree of correlation in hidden activations. In essence, overfitting and co-adaptation
seem to be correlated. The open question of course is -- is the relationship causal? 

This leads to the principal questions of this paper -- 
Is it possible to bias networks towards decorrelated representations by directly 
reducing correlation between hidden units? And do such decorrelated representations generalize better? 

\textbf{Overview and Contributions.}
The goal of this paper is to learn DNNs with decorrelated activations and 
study the effect of this decorrelation on their generalization performance.
Towards this end, we propose a fairly natural
loss called \DeCov, which explicitly
encourages decorrelation between the activations in a deep neural network. 
This loss requires no additional supervision, so it can be added to any existing network. 

In addition to the link discussed above, our motivation also comes 
from the classical literature on bagging and ensemble averaging~\citep{hansen1990neural, Perrone93whennetworks, Breiman1996},
which suggests that decorrelated ensembles perform better than correlated ones. 

Our experiments encompass a range of datasets (MNIST~\citep{lenet}, CIFAR10/100~\citep{cifar10}, 
ImageNet~\citep{imagenet}), 
and different kinds of network architectures (Caffe implementations of
LeNet~\citep{lenet}, AlexNet~\citep{krizhevsky_nips12}, and Network in Network~\citep{lin2013nin}).
All cases suggest that \DeCov acts as a novel and useful regularizer.

\vspace{\sectionReduceTop}
\section{Approach: \DeCov Loss} \label{sec:approach}
\vspace{\sectionReduceBot}

To express our notion of redundant or co-adapated features, we impose a loss on the activations 
of a chosen hidden layer. 
In a manner similar to Dropout, our proposed Decov loss may be applied to a single layer or multiple
layers in a network. For simplicity, let us focus on a single layer.
Let $\mathbf{h}^n \in \mathbb{R}^d$ 
denote the activations at the chosen hidden layer, 
where $n \in \{1, \ldots, N\}$ indexes one example from a batch of size $N$.
The covariances between all pairs of activations
$i$ and $j$ form a matrix $C$:
\beqn
    C_{i, j} = \frac{1}{N} \sum_n (h_i^n - \mu_i) (h_j^n - \mu_j)
\eeqn
where $\mu_i = \frac{1}{N}\sum_n h_i^n$ is the sample mean of activation $i$ over the batch.

We want to minimize covariance between different features, which corresponds to penalizing 
the norm of $C$. However, the diagonal of $C$ contains the variance of each hidden 
activation and we have no reason to require the dynamic range of activations 
to be small, so we subtract this term from the matrix norm to get our final \DeCov loss
\beqn
    \mathcal{L}_{\DeCov} = \frac{1}{2} \left( \| C \|_F^2 -  \| diag(C) \|_2^2 \right)
\eeqn
where $\| \cdot \|_F$ is the frobenius norm, and the $diag(\cdot)$ operator extracts the main diagonal 
of a matrix into a vector. 
In our experiments, subtracting the diagonal made little difference for small 
networks, but led to increased stability for larger networks. 

Perhaps the best quality of this loss is that it requires no supervision, so 
it can be added to any set of activations.
In a manner similar to Dropout, our experiments typically apply Decov loss to
fully connected layers towards the deep end of a network (\eg, fc6 and fc7 for AlexNet).
However, note that Decov affects \emph{all parameters} up to the layer where it
is applied (and not just the parameters in the specific layer). 

At first glance, one seeming peculiarity about this loss is that its global minimum can be found 
by setting all weights for $\mathbf{h}$ to 0. This is similar to an 
L$_2$ regularizer in that both encourage weights to tend toward 0, but
one important difference between these two regularizers is that
$\mathcal{L}_{\DeCov}$ depends on input data and is not a function
purely of a weight vector like one might find in a classical regularizer such as $L_2$ or $L_1$.

To understand this further, consider the gradient
of the loss with respect to a particular
activation $a$ for a particular example $m$
\beqn
    \frac{\partial \mathcal{L}_{\DeCov}}{\partial h_a^m} =
\frac{1}{N} \sum_{j \ne a} \left[ \frac{1}{N} \sum_n (h_a^n - \mu_a) (h_j^n - \mu_j) \right] (h_j^m - \mu_j).
\eeqn
Let us denote the rightmost term in this expression by $I(j, m) = (h_j^m - \mu_j)$.

This term is large (in absolute value) when feature $j$
is discriminative for example $m$ \wrt the mean of the batch.
If $j$ were not discriminative for $m$ then
$h_j^m$ would be close to $\mu_j$.
Hence, we can consider $I$ as an ``importance'' term, corresponding to 
a notion of how significant feature $j$ is for example $m$.

Also notice that the term on the left in the gradient expression is simply the covariance 
between feature $a$ and feature $j$. 
Thus, the gradient can be re-written as 
\beqn
    \frac{\partial \mathcal{L}_{\DeCov}}{\partial h_a^m} = \frac{1}{N} \sum_{j \ne a} C_{a, j} \cdot I(j, m).
\eeqn

\textbf{Interpretation.} 
Intuitively, the covariance term can be thought of as measuring (linear) redundancy: 
features $a$ and $j$ are redundant if they vary together. 
Thus, the \DeCov loss tries to prevent features from being redundant, 
but redundancy is weighted by importance ($I$). 
Specifically, a feature $j$ contributes towards a large gradient of feature $a$ on example $m$ 
if $j$ is important for $m$ \emph{and} correlated with $a$. 
This means important features correlated with $a$ (\eg, $j$) contribute to a large
gradient of $a$, suppressing the activation $h_a^m$. 
A feature which fires only in specialized situations (\eg, a cat's ear) will likely 
be nearly identical or noisy for most other examples (\eg, non-cats) and will not contribute towards 
gradients of other specialized features.

\vspace{\sectionReduceTop}
\section{Related Work}
\vspace{\sectionReduceBot}

\paragraph{Redundancy Based Representations.}
The idea of using low redundancy to learn representations
has been around for decades. In an early attempt to
model human perception, \cite{barlow1961} lists
3 possible learning principles, the 3rd being the
notion that representations should not be redundant.

Later work continued to investigate this intuition in the context
of unsupervised feature learning.
Three objectives emerged, each of which formalize the notion differently.
(1) An information theoretic view is expressed
by~\cite{linsker1988}. The main idea is to maximize information gained
by predicting the next representation/layer between input and output.
(2)
The closest objective to ours is cross-correlation (not cross-covariance), which appears
in~\citep{bengio2009slow} and complements a temporal coherence objective.
It also appears in~\citep{pearlmutter1986gmax} where it complements an objective which encourages
units to capture higher order input statistics.
(3)
Finally, redundancy minimization is realized through
predictability minimization in~\citep{schmidhuber1992} for the purpose
of learning factorial codes (representations whose units are independent).
This objective says that one unit should not be predictable
given \emph{all} of the others in its layer as input.

All of these works focus on unsupervised feature learning and do not experiment
with supervised models. Furthermore, these early pioneering works were
limited by data and evaluated small networks without many of the
modern design choices and features (e.g. ReLus, Dropout, SGD instead of Hebb's update
rule, batch-normalization, \etc).
We propose redundancy minimization for a new purpose (regularization),
evaluate it using modern techniques such as end-to-end learning using SGD
with respect to a supervised objective, and do this in the context of
harder challenges presented by modern datasets.
To the best of our knowledge, such a setting has not been considered before.

\paragraph{Correlation/Covariance Losses in Other Settings.}
Other works have used similar penalties, but in different settings and to different effects.
Deep Canonical Correlation Analysis (Deep CCA)~\citep{andrew2013deep}
and Correlational Neural Networks (CorrNets)~\citep{chandar2015corrnet} apply a similar
loss which \emph{maximizes} correlation, unlike our \emph{minimization} of cross-covariance.
Both methods are used to learn better features in the presence of multiple
views or modalities.
They embed inputs to a common space and maximize correlation between
aligned pairs.

Another idea similar to ours is that of~\cite{hidden_factors}, 
which aims to discover and disentangle hidden factors.
The goal is to separate supervised factors of variation (\eg, class of MNIST digits) 
from unsupervised factors of variation (\eg, handwriting style). 
In order to achieve this goal, they impose a covariance (not correlation) loss 
between (1) the softmax outputs of a neural network trained to recognize digits and 
(2) a hidden representation which is used in conjunction with (1) to reconstruct
the input (via an auto-encoder). 

These two works suggest that correlation losses significantly 
impact learned representations in the context of modern networks.
One key difference between these two approaches and ours is that 
while their formulations decorrelate~\citep{hidden_factors} and disregard~\citep{andrew2013deep,chandar2015corrnet} parts 
of \emph{different} representations, our approach tries to decorrelate parts of the 
\emph{same} representation. 
Moreover, the ultimate goals are different. Unlike these approaches, 
our goal is simply to improve supervised classification performance by reducing overfitting, 
and not to reconstruct the original data. 

\textbf{Dropout and Batch Normalization.}
Two recent approaches to regularization 
in deep neural networks are Dropout~\citep{dropout} and to some extent
Batch Normalization~\citep{ioffe2015batch}. 
Dropout aligns with our intuition and goals more closely as it
aims to improve classification performance by reducing co-adaptation
of activations.
On the other hand, Batch Normalization focuses on faster optimization by
reducing \emph{internal co-variate shift}, which is the constant variation
of a layer's input as it learns. Some Batch Normalization results indicate it could
act as a regularizer, but this
has not been exhaustively verified yet.
Our approach is similar to Batch Normalization
due to its use of mini-batch statistics.

\vspace{\sectionReduceTop}
\section{Experiments}
\vspace{\sectionReduceBot}

We begin with a synthetic dual ``modality'' experiment, which
serves as a testbed
for measuring improvement due to decorrelation.
Next, we use an autoencoder (as in \cite{dropout})
to contrast \DeCov and Dropout. Finally, we use a variety of experiments to report 
Image Classification performance on CIFAR10/100 and ImageNet, noticing
significant improvement in \emph{all cases}.
Note that we set the Dropout rate to 0.5 as suggested by ~\cite{dropout}.

\subsection{Dual modality experiments with MNIST: Predicting Side-by-Side Digits}
\label{sec:mnist_bias}

\begin{figure}[h]
    \centering
    \includegraphics[scale=0.4]{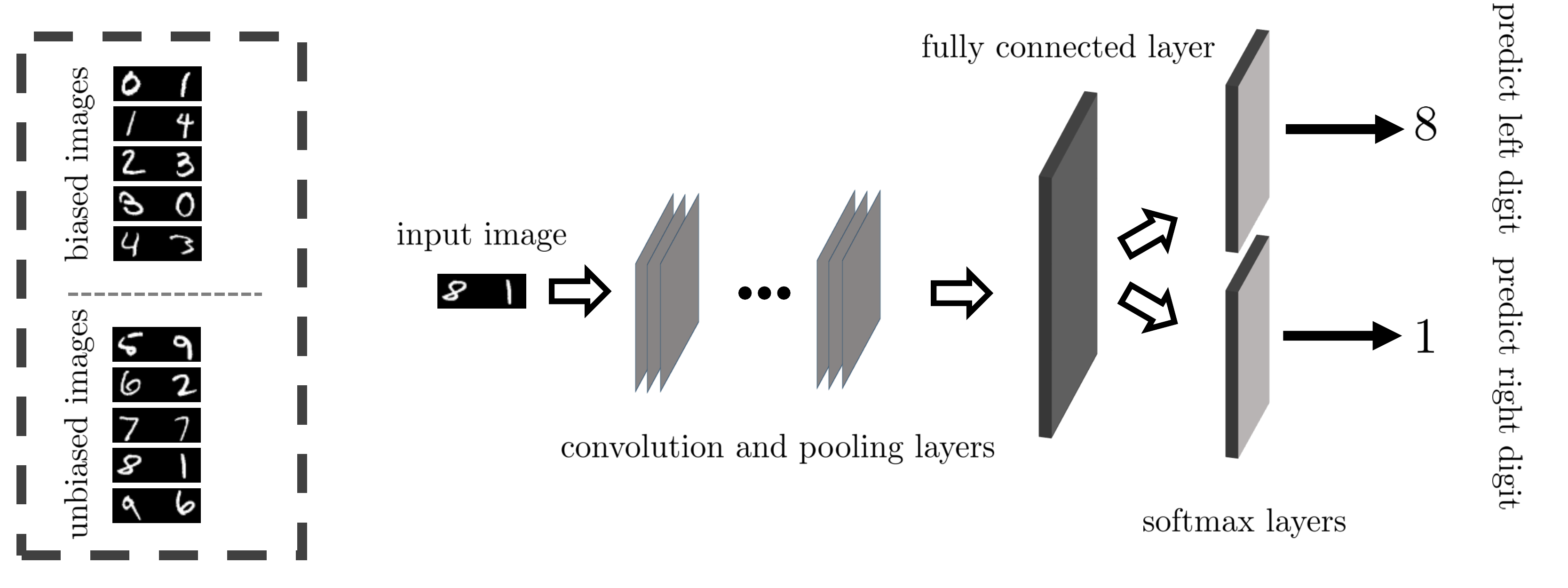}
    \caption{We consider the task of simultaneously predicting two MNIST digits placed side by side.
    By biasing right digits more than left digits at train time, we create a
    controlled scenario with the type of problem we expect \DeCov to solve.}
    \vspace{-5pt}
    \label{fig:sbs}
\end{figure}

We propose a synthetic dual ``modality'' task on MNIST --  simultaneously predict the class labels 
for two digits placed adjacent in an image.
We created a dataset where each example consists
of two MNIST digit images horizontally concatenated and
separated by 16 black pixels (to prevent interference between feature maps in the first layers). 
\figref{fig:sbs} shows a few examples. 

The important detail of this experiment is the particular bias we inject into the distribution
of left and right digits. Let
\beqn
P(l) = 0.1 \;\; \text{and} \;\; P(r | l) =
    \begin{cases}
        0   & \text{if } l \in \{0, \ldots, 4\} \text{ and } r \in \{0, \ldots, 4\} \\
        0.2 & \text{if } l \in \{0, \ldots, 4\} \text{ and } r \in \{5, \ldots, 9\} \\
        0.1 & \text{if } l \in \{5, \ldots, 9\} \\
    \end{cases}.
\eeqn
To generate one example we first sample the left digit using $P(l)$ then the right using $P(r | l)$.
As shown in Appendix \ref{appendix:mnist_bias}, we can compute the conditional entropies
of one digit given the other to get $H(l | r) = 2.0868$ and $H(r | l) = 1.9360$. Since
$H(l | r) > H(r | l)$, the left digit is more informative of the right than the right is
of the left. There is no cross-digit signal at test time, so features for the right and
left digits should be completely decorrelated to generalize, but learned features
will have some correlation between left and right. Intuitively, \DeCov should help
generalization in this scenario. Our experiments support this.

We use Caffe's~\citep{caffe} reference version of LeNet~\citep{lenet}. 
It has two convolution layers, each followed by pooling, then a fully connected 
layer with 500 hidden units which are shared between the two softmax layers. 
We apply \DeCov and/or Dropout to the 500 hidden units of the fully connected layer.

\begin{table}[h]
\vspace{10pt}
\small
\setlength{\tabcolsep}{4pt}
\begin{center}
\begin{tabular}{@{}  l  l  c  c  c  c  c  c  @{}}
\toprule
& & \multicolumn{3}{c}{Left Digit} & \multicolumn{3}{c}{Right Digit} \\
\cmidrule[0.75pt](l){3-5}
\cmidrule[0.75pt](lr){6-8}
\DeCov & Dropout & train & test & train - test & train & test & train - test \\
\midrule
no  & no  & 99.98 $\pm$ 0.01 & 97.94 $\pm$ 0.18 & 2.05 $\pm$ 0.19 & 100.00 $\pm$ 0.00 & 96.75 $\pm$ 0.24 & 3.25 $\pm$ 0.24 \\
no  & yes & 99.99 $\pm$ 0.00 & 98.45 $\pm$ 0.04 & 1.54 $\pm$ 0.04 & 99.99 $\pm$ 0.00 & 97.39 $\pm$ 0.20 & 2.61 $\pm$ 0.20 \\
yes & yes & 99.97 $\pm$ 0.01 & 98.59 $\pm$ 0.12 & 1.38 $\pm$ 0.12 & 99.99 $\pm$ 0.00 & 97.81 $\pm$ 0.07 & 2.18 $\pm$ 0.06 \\
yes & no  & 99.99 $\pm$ 0.00 & \textbf{98.74} $\pm$ \textbf{0.03} & \textbf{1.25} $\pm$ \textbf{0.04} & 99.99 $\pm$ 0.00 & \textbf{97.99} $\pm$ \textbf{0.12} & \textbf{2.00} $\pm$ \textbf{0.12} \\
\midrule
\multicolumn{2}{l}{weight decay} & 99.97 & 97.86 & 2.11 & 99.97 & 96.21 & 3.76 \\
\bottomrule
\end{tabular}
\caption{\textbf{MNIST side by side results.}
As expected, biasing right digits at train time so that they are weakly informed
by left digits leads to lower performance on an unbiased test set.
More importantly, \DeCov provides greater improvements
over the baselines on the right, confirming that it leads
to better features when decorrelation is extremely likely to improve performance.
}
\label{tab:mnist_sbs}
\end{center}
\end{table}

\textbf{Results.} 
\tableref{tab:mnist_sbs} reports the accuracy of left and right digit
classifiers. Our injected dataset bias can be clearly seen in the lower test accuracy
and higher train-test gap of the right classifier, indicating that all
of our networks incorporate the train time bias into their predictions.
We report mean accuracies across 4 trials, along with the standard 
deviation. We also compare the effect of Dropout. 

The main result is that the gaps between the performance of \DeCov and the baselines
are larger for the biased right digit (\eg, right digit test accuracy shows a $\sim$0.6\%
improvement when switching from Dropout-alone to \DeCov-alone
while the improvement for left digits is just $\sim$0.3\%).
This suggests that the baselines pick up
on the false bias and that \DeCov does the best job of correcting for it.
\DeCov also improves generalization for both classifiers since test accuracy is higher
in the bottom two rows and the train - test gap is lower in those rows.
Combining Dropout with our \DeCov loss
hurts slightly, but we note that the error bars overlap in some cases,
so this is not a statistically significant difference. 

One skeptical hypothesis is that the \DeCov loss is simply enforcing something akin to
an L2 penalty on the weights. The experiments with \DeCov and Dropout already
use an L2 penalty, so this is unlikely, but a grid search over weights on
this term shows it makes little difference.
The best accuracies are reported in the last row of \tableref{tab:mnist_sbs}.

\subsection{MNIST Autoencoder}

\begin{figure}[t]
    \centering
	\begin{subfigure}[b]{0.30\textwidth}
        \includegraphics[width=\textwidth]{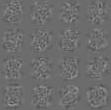}
        \caption{Baseline with train MSE = 1.47 and test MSE = 1.47}
        \label{fig:auto_base}
    \end{subfigure}
	\begin{subfigure}[b]{0.30\textwidth}
        \includegraphics[width=\textwidth]{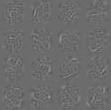}
        \caption{\DeCov with train MSE = 0.98 and test MSE = .98}
        \label{fig:auto_decov}
    \end{subfigure}
	\begin{subfigure}[b]{0.30\textwidth}
        \includegraphics[width=\textwidth]{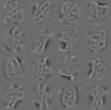}
        \caption{Dropout with train MSE = 3.08 and test MSE = 3.03}
        \label{fig:auto_drop}
    \end{subfigure}
    \caption{Weights learned by the first layer of a 2 layer autoencoder are
        reshaped into images and visualized for a model with no \DeCov or Dropout (\figref{fig:auto_base}),
        a model with \DeCov (\figref{fig:auto_decov}), and a model with Dropout (\figref{fig:auto_drop}).}
    \vspace{-10pt}
    \label{fig:autoencoder}
\end{figure}

To offer a more qualitative point of comparison, we visualized learned features using
the 2 layer autoencoder experiment
from~\citep{dropout} (section 7). In this experiment an autoencoder
is trained on raw pixels of single MNIST digits
using an encoder with 1 layer of 256 ReLU units and a decoder (untied weights)
that produces 784 ($28 \times 28$) ReLU outputs.
\figref{fig:autoencoder} shows the weights learned by the autoencoder (reshaped to align with
the input image) and mean-square reconstruction errors. 

Weight initialization turned out to be an important factor for the visualizations.
Initializing all weights by sampling from $U[-\sqrt\frac{3}{n}, \sqrt\frac{3}{n}]$
(based on~\cite{glorot2010understanding}; as implemented in Caffe) led to visualizations as seen in
\cite{dropout} (the baseline looks like noise), but sampling weights from a
Gaussian with mean 0 and standard deviation 0.001 led to baseline
visualizations with faint digit outlines. The latter initialization was
used in \figref{fig:autoencoder}.

One take-away is that MSE is significantly lower for \DeCov than others.
However, the key take-away is the qualitative difference between representations learned
with Dropout and those learned with \DeCov.
Recall from \secref{sec:intro} that Dropout reduces cross-covariance while \DeCov
explicitly minimizes it. Despite this intuitive similarity, the two lead to
different learned representations. 

\subsection{Image Classification}

\subsubsection{CIFAR10} 
CIFAR10 contains 60,000 32x32
images sorted into 10 distinct categories \citep{cifar10}.
We training on the 50,000 given training
examples and testing on the 10,000 specified test samples. 
Hyper-parameters (loss weights for \DeCov and weight decay) are chosen by
grid search on the standard train/val split. 

We use Caffe's quick CIFAR10 architecture, which has 3 convolutional layers 
followed by a fully connected layer with 64 hidden units and a softmax layer.
The hidden fully connected layer is not followed by a non-linearity.
The \DeCov loss is added only to the 64 hidden units in the hidden fully connected layer.
All reported results are average performance over 4 trials with the standard deviation indicated alongside.

\begin{table}[h]
\small
\setlength{\tabcolsep}{13pt}
\begin{center}
\begin{tabular}{@{}  l  l  c  c  c  @{}}
\toprule
\DeCov & Dropout & train & test & train - test \\
\midrule
no  & no  & 100.0 $\pm$ 0.00 & 75.24 $\pm$ 0.27 & 24.77 $\pm$ 0.27 \\
no  & yes & 99.10 $\pm$ 0.17 & 77.45 $\pm$ 0.21 & 21.65 $\pm$ 0.22 \\
yes & yes & 87.78 $\pm$ 0.08 & \textbf{79.75} $\pm$ \textbf{0.17} & \textbf{8.04} $\pm$ \textbf{0.16} \\
yes & no  & 88.78 $\pm$ 0.23 & 79.72 $\pm$ 0.14 & 9.06 $\pm$ 0.22 \\
\midrule
\multicolumn{2}{c}{weight decay} & 100.0 & 75.29 & 24.71 \\
\bottomrule
\end{tabular}
\caption{CIFAR10 Classification.
We can see that \DeCov with Dropout leads to the highest test performance and the lowest train-test gap.}
\label{tab:cifar10}
\end{center}
\end{table}

\paragraph{Results.} 
In Table \ref{tab:cifar10}, we again
observe significant improvements when using the \DeCov loss -- 
there is a $\sim$4.5\% improvement in test accuracy (over no regularization).
Moreover, the \DeCov loss reduces the gap between train and val accuracies by $\sim$15\% (without Dropout)
and $\sim$16\% (with Dropout)! 

Comparing the four combinations, we see that using \DeCov alone provides a larger 
improvement than using Dropout. Using both \DeCov 
and Dropout further improves the generalization (as measured by the gap in train and test 
accuracies), but the improvement in absolute test performance does not seem statistically significant. 

We again test if L2 weight decay can provide similar improvements and 
find once again that the best setting gives little improvement over the baseline.

One promise of regularization is the ability to train larger networks, 
so we increase the size of our CIFAR10 network. We 
add another fully connected layer to the network used in the previous experiment,
double the number of filters in each convolutional layer, and double the 
number of units in the fully connected layers. This larger network performs better
than the smaller version -- all accuracies are higher than corresponding entries in 
\tableref{tab:cifar10}. However, there are the stronger indications of 
overfitting in this network -- specifically, the train accuracies are much higher than test accuracies 
(when compared to the previous network). 
\tableref{tab:cifar10_bigger} shows the results. 
We observe similar trends as the previous experiment -- 
there are significant gains from using \DeCov alone compared to Dropout alone, 
and there is a further slight improvement in combining both. 
Using Dropout alone gives a $\sim$1.5\% boost in test accuracy, while using \DeCov alone 
provides a $\sim$4\% increase in test accuracy. 
Using both yields roughly the same test performance, but the trainval and test gap is further reduced. 

\begin{table}[h]
\small
\setlength{\tabcolsep}{5.5pt}
\begin{center}
\begin{tabular}{@{}  l  l  c  c  c  @{}}
\toprule
\DeCov & Dropout & (train+val) & test & (train+val) - test \\
\midrule
no  & no  & 100.00 & 77.38 & 22.62 \\
no  & yes & 100.00 & 79.93 & 20.07 \\
yes & yes & 96.76 & \textbf{81.68} & \textbf{15.08} \\
yes & no  & 98.15 & 81.63 & 16.52 \\
\bottomrule
\end{tabular}
\caption{CIFAR10 Classification with a bigger version of the base network}
\label{tab:cifar10_bigger}
\end{center}
\end{table}

\subsubsection{CIFAR100} 
To scale up our experiments, we move to CIFAR100~\citep{cifar10}. We use the 
same architecture as the base architecture for CIFAR10 and hold out the
last 10,000 of the 50,000 train examples for validation.
Table \ref{tab:cifar100} 
shows that Dropout alone highest higher test performance than \DeCov alone, 
but \DeCov leads to a smaller train-test gap. 
Using both regularizers not only achieves the highest test accuracy, but also the smallest 
train-test gap ($\sim$34\% smaller than using neither regularizer).
This suggests that the two regularizers may have complementary effects.

\begin{table}[h]
\small
\setlength{\tabcolsep}{9pt}
\begin{center}
\begin{tabular}{@{}  l  l  c  c  c  @{}}
\toprule
\DeCov & Dropout & train & test & train - test \\
\midrule
no  & no  & 99.77 & 38.52 & 61.25 \\
no  & yes & 87.35 & 43.55 & 43.80 \\
yes & yes & 72.53 & \textbf{45.10} & \textbf{27.43} \\
yes & no  & 77.92 & 40.34 & 37.58 \\
\bottomrule
\end{tabular}
\caption{CIFAR100 Classification Accuracies}
\label{tab:cifar100}
\end{center}
\end{table}

One more problem comes with the question of 
how to weight the \DeCov loss. All of our experiments use grid search to pick this hyper-parameter.
The optimal weight varies across datasets, but 
we have found consistency across variations in architecture.
We varied both the \DeCov weight and the number of hidden units in the
fully connected layer to which \DeCov is applied, training a new network for
each setting.
The best \DeCov weight (0.1)
is consistent for a range of hidden activation sizes in this dataset,
though it is different in other experiments.

\subsubsection{ImageNet} 

\begin{figure*}[h]
        \vspace{-5pt}
        \includegraphics[scale=0.5]{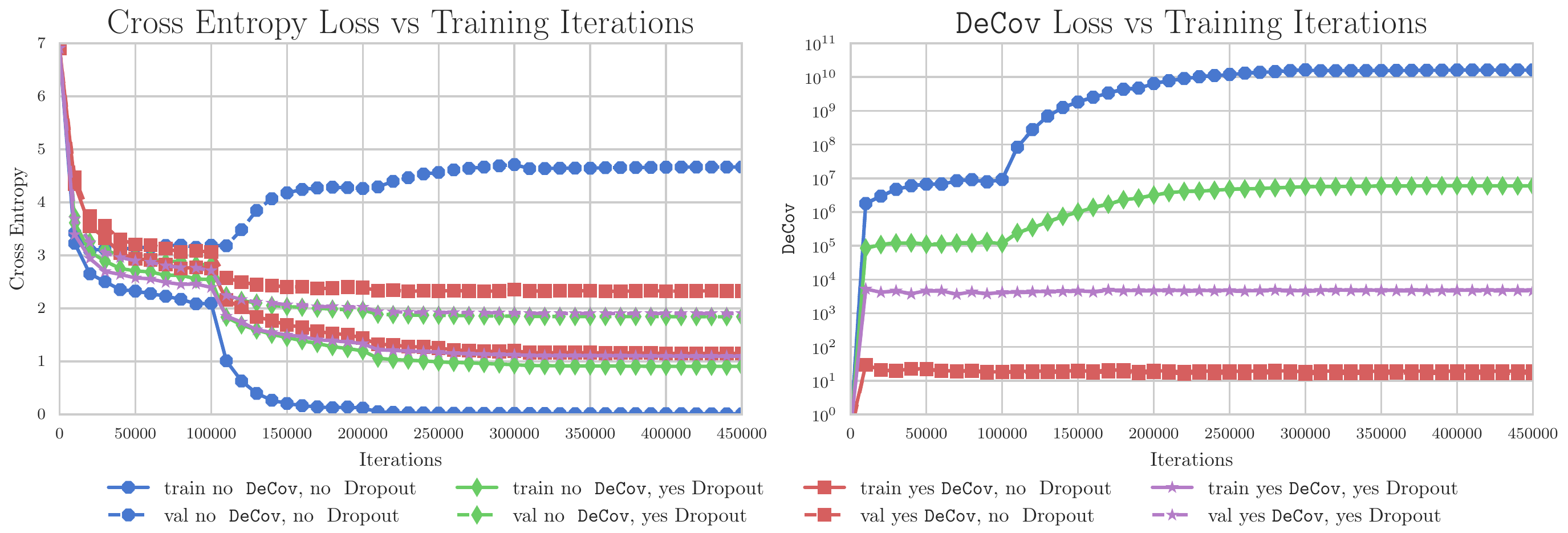}
        \centering
        \vspace{1pt}
        \caption{Cross Entropy and \DeCov losses over the course of training AlexNet with 256x256 images. Note that the \DeCov val
        curves are hidden by the train curves.
        Interestingly, \DeCov is reduced even by Dropout, though not nearly as much as when it is explicitly minimized.}
                \label{fig:imagenet_losses}
\end{figure*}

Now we explore results for networks trained for ImageNet classification, starting by applying \DeCov to fc6 and fc7 in AlexNet~\citep{krizhevsky_nips12}.
The last 50,000 of the ILSVRC 2012 train images are held out for validation.
Our implementation comes from Caffe. In particular, it uses a fixed schedule that multiplies the learning
rate by 1/10 every 100,000 iterations (see jumps in \figref{fig:imagenet_losses}). We do not use early stopping and do not perform color augmentation.

In \figref{fig:imagenet_losses} we notice that when neither of the two regularizers -- Dropout or \DeCov -- are applied (blue line), the network overfits (it even gets 100\% train accuracy), and
the \DeCov loss (hidden activation redundancy) is higher than with any other combination of the regularizers. Applying either of the regularizers also causes a synchronous drop in both losses.
Explicitly minimizing the \DeCov loss naturally leads to much lower \DeCov losses, and we notice that this coincides with
significantly reduced overfitting. Interestingly, Dropout results in relatively lower \DeCov loss too, even when \DeCov is not optimized
for. This is further indication of the link between redundant activations and overfitting.

\begin{figure*}[h]
        \includegraphics[scale=0.5]{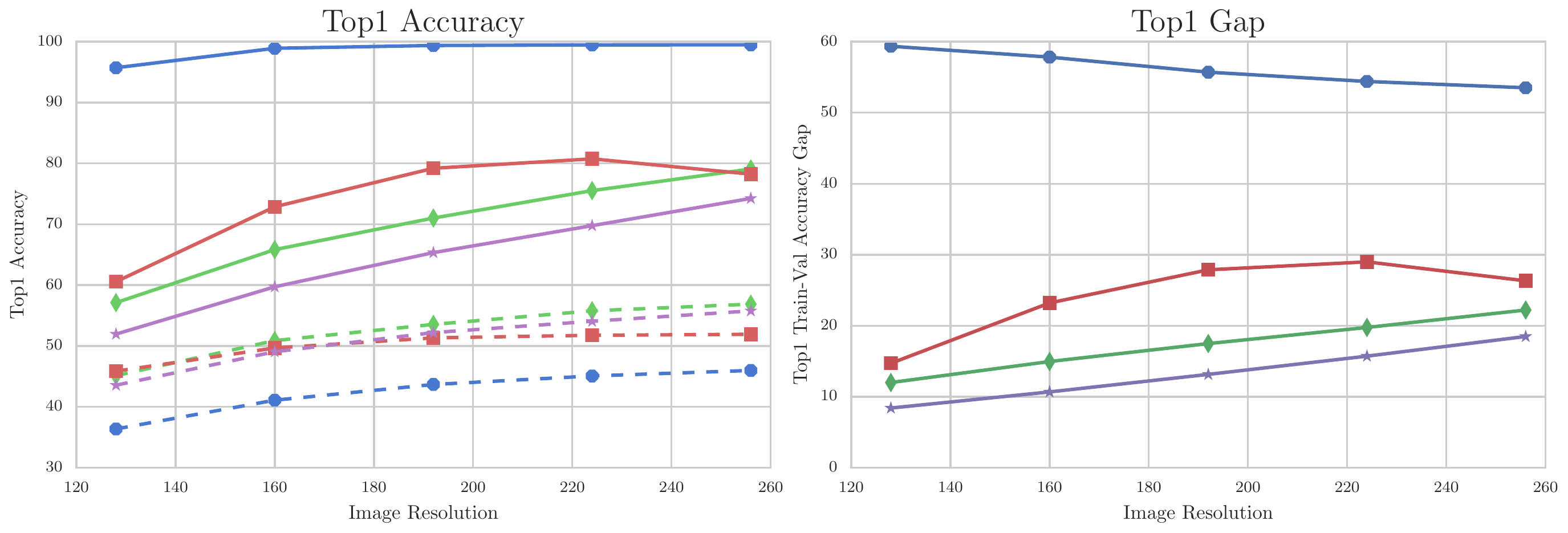}
        \includegraphics[scale=0.5]{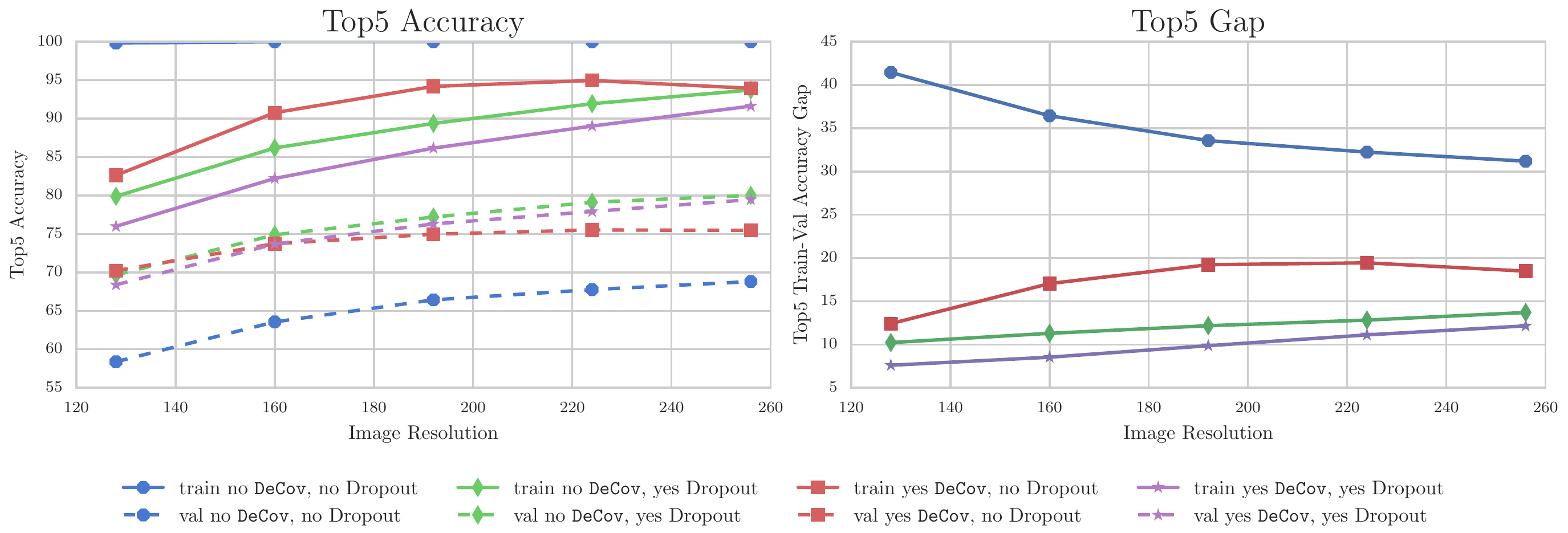}
        \centering
        \vspace{1pt}
\caption{ImageNet classification performance using AlexNet. Plots on the left show training and validation (ILSVRC 2012 validation set)
accuracy at different resolutions.
Note how all curves have a much lower train-val gap than the (blue) baseline.}
\label{fig:alexnet}
\end{figure*}

\figref{fig:alexnet} shows accuracies across different image resolutions we used to train AlexNet.
AlexNet is typically trained with 256x256 images, but training with smaller images is faster
\footnote{
Using CuDNNv3, AlexNet with 128x128 inputs takes 103ms averaged over 50 runs to compute a forward and backward pass.
For 256x256 images this time is 449ms.}
\emph{and}
reduces the number of parameters in the network. Smaller images (we use 128x128, 160x160, 192x192, and 224x224) lead to smaller feature maps output by pool5, so
the dense connection between pool5 and fc6 has fewer parameters, the model has less
capacity, and it's less likely to overfit. For example, images scaled to 256x256 (taking 227x227 crops
\footnote{At train time crops are sampled and mirrored randomly. At test time only the center 227x227 crop is used.})
lead to a weight matrix with 38 million parameters while 128x128 images (with 99x99 crops)
result in a 4 million parameter matrix.
Generally, accuracies (left plots) and the train-val gap (right plots) have a slight positve slope,
confirming that performance and overfitting increase with resolution and model capacity.
Note that the \DeCov loss weight was tuned using grid search at each resolution both with and without Dropout.

We see that Dropout alone (green) usually has the best val accuracy, which is slightly
higher than the two losses combined (purple) and a couple points higher than \DeCov alone (red)
at higher resolutions. At the lowest resolution Dropout alone is tied with \DeCov alone.
Dropout also reduces overfitting more than \DeCov, though both independently reduce 
overfitting by a large margin -- from 59.35\% to 14.7\% in the case of \DeCov @ 128x128.

Finally, we test our new regularizer on ILSVRC 2012 with one more architecture -- 
the Network in Network \citep{lin2013nin}.\footnote{This is the model provided in the Caffe Model Zoo: \url{https://gist.github.com/mavenlin/d802a5849de39225bcc6}} 
This architecture is fully convolutional: it contains 4 convolutional layers,
with 96, 256, 384, and 1024 feature maps, respectively. Between each of these
layers and after the last are two convolutional layers which have 1x1 kernels,
which further process each feature map output by the main convolutional layers
before being fed into the next layer. To produce 1000 softmax activations, 
1000 feature maps are averaged over spatial locations to produce one feature 
vector. We applied \DeCov to these average pooled feature vectors. 

Interestingly, this architecture has much less overfitting than AlexNet.
However, adding a \DeCov loss still decreases overfitting substantially 
and improves validation accuracy. There is a small boost in performance on
validation accuracies and a significant decrease of $\sim$3\% (for top 1)
and $\sim$2\% (for top 5) in the train - val gap.

\begin{table}[h]
\small
\setlength{\tabcolsep}{5.5pt}
\begin{center}
\begin{tabular}{@{}  l  l  c  c  c  @{}}
\toprule
\DeCov & Dropout & ILSVRC 2012 train top 1 & ILSVRC 2012 val top 1 & train - val \\
\midrule
no  & no  & 71.68 & 58.67 & 13.01 \\
no  & yes & 71.32 & 58.95 & 12.37 \\
yes & yes & 68.28 & \textbf{59.08} & \textbf{9.20}  \\
yes & no  & 68.33 & 58.85 & 9.48  \\
\midrule
\DeCov & Dropout & ILSVRC 2012 train top 5 & ILSVRC 2012 val top 5 & train - val \\
\midrule
no  & no  & 89.91 & 81.18 & 8.73 \\
no  & yes & 89.63 & 81.53 & 8.10 \\
yes & yes & 87.99 & \textbf{81.94} & \textbf{6.05} \\
yes & no  & 87.88 & 81.57 & \textbf{6.05} \\
\bottomrule
\end{tabular}
\caption{ImageNet Classification Accuracies with Network in Network}
\label{tab:nin}
\end{center}
\end{table}

\vspace{\sectionReduceTop}
\section{Discussion and Conclusion}
\vspace{\sectionReduceBot}

\textbf{Fine-tuning.}
In the experiments we presented, networks were always trained from scratch,
but we also tried fine-tuning networks in different scenarios. During our 
ImageNet experiments we fine-tuned both the Network in Network and AlexNet
architectures initialized with parameters that weren't trained with a \DeCov loss,
but were trained with Dropout. In both cases performance either stayed where it
was at fine-tuning initialization or it decreased slightly (within statistical significance). 
We found similar results when fine-tuning for other tasks like attribute classification (fine-tuning 
AlexNet) and object detection (Fast RCNN~\citep{girshick2015fast}).

This, along with some cases where combining Dropout and \DeCov decreases 
performance slightly suggest that the \DeCov loss may possibly be acting adversarially to activations 
learned by Dropout. 
Fine-tuning with \DeCov is an interesting direction for future work.

\textbf{Trends. }
All of our experiments strongly indicate two clear trends:
\begin{compactenum}
\item
\DeCov reduces overfitting as measured by the gap between train and test performance.

\item
\DeCov acts as a regularizer: performance with \DeCov is always better than
performance without either \DeCov or Dropout.
\end{compactenum}

To be clear, the results do not support that Dropout can be completely replaced by DeCov, but simply that in a number of scenarios DeCov is a useful alternative and their combination almost always works the best. 
Our loss clearly has desirable regularization
properties at the expense of one extra hyper-parameter to tune.

In this work, we proposed a new \DeCov loss which explicitly penalizes the covariance between the 
activations in the same layer of a neural network in an unsupervised fashion.
This loss acts as a strong
regularizer for deep neural networks, where overfitting
is a major problem and Dropout has been required to get large models to generalize
well. We show that \DeCov competes well against Dropout over a range
of experiments which investigate different scales, datasets and architectures.

\textbf{Acknowledgements. }
This work was supported in part by the following awards to DB: National Science Foundation CAREER award, Army Research Office YIP award, Office of Naval Research grant N00014-14-1-0679, AWS in Education Research Grant, and GPU support by NVIDIA. The views and conclusions contained herein are those of the authors and should not be interpreted as necessarily representing the official policies or endorsements, either expressed or implied, of the U.S. Government or any sponsor.



{
\footnotesize
\bibliographystyle{iclr2016_conference}
\bibliography{dbatra}
}


\appendix
\section*{Appendices}
\addcontentsline{toc}{section}{Appendices}
\renewcommand{\thesubsection}{\Alph{subsection}}
\subsection{Details of the bias in the MNIST experiment} \label{appendix:mnist_bias}

Recall that in \secref{sec:mnist_bias} we generate biased pairs of MNIST digits by defining
\beqn
P(l) = 0.1 \;\; \text{and} \;\; P(r | l) =
    \begin{cases}
        0   & \text{if } l \in \{0, \ldots, 4\} \text{ and } r \in \{0, \ldots, 4\} \\
        0.2 & \text{if } l \in \{0, \ldots, 4\} \text{ and } r \in \{5, \ldots, 9\} \\
        0.1 & \text{if } l \in \{5, \ldots, 9\} \\
    \end{cases}
\eeqn
and sampling left then right digits. To show that this creates a larger bias on the right
than on the left, we show there is more uncertainty about left digits given right ones than
right ones given left ones. That is, we show the conditional entropy $H(l | r)$ is greater
than $H(r | l)$.

To compute the conditional entropies, we first derive
\beqn
    P(r) = \sum_l P(r | l) P(l) =
    \begin{cases}
        0.05 & \text{if } r \in \{0, \ldots, 4\} \\
        0.15 & \text{if } r \in \{5, \ldots, 9\} \\
    \end{cases}
\eeqn
and
\beqn
    P(l | r) = \frac{P(r | l) P(l)}{P(r)} =
    \begin{cases}
        0            & \text{if } l \in \{0, \ldots, 4\} \text{ and } r \in \{0, \ldots, 4\} \\
        \frac{2}{15} & \text{if } l \in \{0, \ldots, 4\} \text{ and } r \in \{5, \ldots, 9\} \\
        \frac{3}{15} & \text{if } l \in \{5, \ldots, 9\} \text{ and } r \in \{0, \ldots, 4\} \\
        \frac{1}{15} & \text{if } l \in \{5, \ldots, 9\} \text{ and } r \in \{5, \ldots, 9\} \\
    \end{cases}.
\eeqn

Using the convention $0 \log{0} = 0$, we can now compute 
\beqn
    H(l | r) = - \sum_r P(r) \sum_l P(l | r) \log{P(l | r)} \approx 2.0868 \\
    H(r | l) = - \sum_l P(l) \sum_r P(r | l) \log{P(r | l)} \approx 1.9560 \\
\eeqn

\end{document}